\documentclass{article}
\usepackage{spconf,amsmath,epsfig}
\usepackage{enumitem}
\usepackage{graphicx}
\usepackage{amsmath}
\usepackage{amssymb}
\usepackage{epsfig}
\usepackage{caption}
\usepackage{booktabs}
\usepackage[export]{adjustbox}
\usepackage{xspace}
\usepackage[table]{xcolor} 
\usepackage{subcaption}
\usepackage{colortbl}
\usepackage{etoolbox}
\usepackage{pgf} 
\usepackage{makecell}
\definecolor{cvprblue}{rgb}{0.21,0.49,0.74}
\usepackage[pagebackref,breaklinks,colorlinks,citecolor=cvprblue]{hyperref}
\usepackage{multirow}
\usepackage{setspace}

\newcommand{\mc}[2]{\multicolumn{#1}{c}{#2}}
\newcommand{\Th}[1]{\textsc{#1}}

\def\l1{\ensuremath{\ell_1}\xspace}
\def\l2{\ensuremath{\ell_2}\xspace}

\newcommand{\cI}{\mathcal{I}}

\newcommand{\cT}{\mathcal{T}}

\makeatletter
\DeclareRobustCommand\onedot{\futurelet\@let@token\@onedot}
\def\@onedot{\ifx\@let@token.\else.\null\fi\xspace}

\makeatother

\newcommand{\ours}{\textsc{WeiCom}\xspace} 
\newcommand{\dataset}{\textsc{PatternCom}\xspace} 

\definecolor{OrangeFrame}{HTML}{D79B00}

\definecolor{textquerycolor}{RGB}{130,179,102}
\definecolor{imagequerycolor}{RGB}{16,115,158}
\definecolor{retrievedimagecolor}{RGB}{150,115,166}

\definecolor{LightSteelBlue1}{RGB}{202, 225, 255}
\definecolor{high}{HTML}{76f013}  
\definecolor{low}{HTML}{ec462e}  
\newcommand*{\opacity}{60}
\newcommand*{\minvalcolor}{14.5}
\newcommand*{\maxvalcolor}{55.3}
\newcommand{\gradientcolor}[1]{
    \pgfmathparse{(#1-\minvalcolor)/(\maxvalcolor-\minvalcolor)}
    \let\normalizedval\pgfmathresult

    \pgfmathparse{100*(\normalizedval)^(2.0)} 
    \xdef\tempa{\pgfmathresult}

    \pgfmathparse{min(100,max(0,\tempa))}
    \xdef\tempa{\pgfmathresult}

    \cellcolor{high!\tempa!low!\opacity} #1
}
\newcommand*{\minvalcontext}{6.6}
\newcommand*{\maxvalcontext}{31.5}
\newcommand{\gradientcontext}[1]{
    \pgfmathparse{(#1-\minvalcontext)/(\maxvalcontext-\minvalcontext)}
    \let\normalizedval\pgfmathresult

    \pgfmathparse{100*(\normalizedval)^(2.0)} 
    \xdef\tempa{\pgfmathresult}

    \pgfmathparse{min(100,max(0,\tempa))}
    \xdef\tempa{\pgfmathresult}

    \cellcolor{high!\tempa!low!\opacity} #1
}
\newcommand*{\minvaldensity}{15.1}
\newcommand*{\maxvaldensity}{42}
\newcommand{\gradientdensity}[1]{
    \pgfmathparse{(#1-\minvaldensity)/(\maxvaldensity-\minvaldensity)}
    \let\normalizedval\pgfmathresult

    \pgfmathparse{100*(\normalizedval)^(3.0)} 
    \xdef\tempa{\pgfmathresult}

    \pgfmathparse{min(100,max(0,\tempa))}
    \xdef\tempa{\pgfmathresult}

    \cellcolor{high!\tempa!low!\opacity} #1
}
\newcommand*{\minvalexistence}{9}
\newcommand*{\maxvalexistence}{16}
\newcommand{\gradientexistence}[1]{
    \pgfmathparse{(#1-\minvalexistence)/(\maxvalexistence-\minvalexistence)}
    \let\normalizedval\pgfmathresult

    \pgfmathparse{100*(\normalizedval)^(0.5)} 
    \xdef\tempa{\pgfmathresult}

    \pgfmathparse{min(100,max(0,\tempa))}
    \xdef\tempa{\pgfmathresult}

    \cellcolor{high!\tempa!low!\opacity} #1
}
\newcommand*{\minvalquantity}{7.0}
\newcommand*{\maxvalquantity}{20.9}
\newcommand{\gradientquantity}[1]{
    \pgfmathparse{(#1-\minvalquantity)/(\maxvalquantity-\minvalquantity)}
    \let\normalizedval\pgfmathresult

    \pgfmathparse{100*(\normalizedval)^(4.0)} 
    \xdef\tempa{\pgfmathresult}

    \pgfmathparse{min(100,max(0,\tempa))}
    \xdef\tempa{\pgfmathresult}

    \cellcolor{high!\tempa!low!\opacity} #1
}
\newcommand*{\minvalshape}{15.2}
\newcommand*{\maxvalshape}{32}
\newcommand{\gradientshape}[1]{
    \pgfmathparse{(#1-\minvalshape)/(\maxvalshape-\minvalshape)}
    \let\normalizedval\pgfmathresult

    \pgfmathparse{100*(\normalizedval)^(8.0)} 
    \xdef\tempa{\pgfmathresult}

    \pgfmathparse{min(100,max(0,\tempa))}
    \xdef\tempa{\pgfmathresult}

    \cellcolor{high!\tempa!low!\opacity} #1
}
\newcommand*{\minvalaverage}{11.3}
\newcommand*{\maxvalaverage}{30.2}

\newcommand{\gradientaverage}[1]{
    \pgfmathparse{(#1-\minvalaverage)/(\maxvalaverage-\minvalaverage)}
    \let\normalizedval\pgfmathresult

    \pgfmathparse{100*(\normalizedval)^(5.0)} 
    \xdef\tempa{\pgfmathresult}

    \pgfmathparse{min(100,max(0,\tempa))}
    \xdef\tempa{\pgfmathresult}

    \cellcolor{high!\tempa!low!\opacity} #1
}

\title{Composed Image Retrieval for Remote Sensing}
\authors
  {Bill Psomas$^{1}$, Ioannis Kakogeorgiou$^{1}$, Nikos Efthymiadis$^{2}$, Giorgos Tolias$^{2}$, \\ Ondřej Chum$^{2}$, Yannis Avrithis$^{3}$, Konstantinos Karantzalos$^{1}$\vspace{0.5em}} 
  {$^1$National Technical University of Athens, $^2$Czech Technical University in Prague, \\ $^3$Institute of Advanced Research in Artificial Intelligence (IARAI)}

\begin{document}
%

\makeatletter
\apptocmd\@maketitle{{\teaser{}}}{}{}
\makeatother
\newcommand{\myfig}[1]{\includegraphics[width=0.125\textwidth,valign=c]{#1}}
\newcommand{\mycaption}[1]{{\rotatebox[origin=c]{0}{\footnotesize#1}}}

\newcommand{\mytextquery}[1]{{\rotatebox[origin=c]{0}{\footnotesize#1}}}
\setlength{\tabcolsep}{5pt}
\newcommand{\boxcolor}{\setlength{\fboxrule}{.8pt} \color{textquerycolor}}
\newcommand{\boxcolorim}{\setlength{\fboxrule}{.8pt} \color{imagequerycolor}}

\newcommand{\teaser}{%
\vspace{-15pt}
\centering
\setlength{\tabcolsep}{2pt}
\begin{tabular}{@{}ccccccc@{}}

 {\boxcolorim \fbox{\myfig{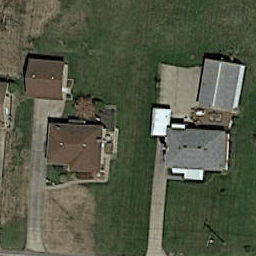}}} &
    \myfig{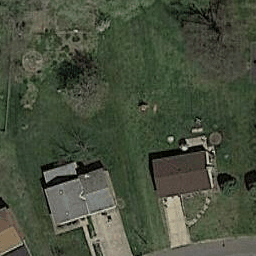} &
    \myfig{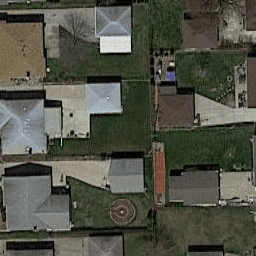} &  
    \myfig{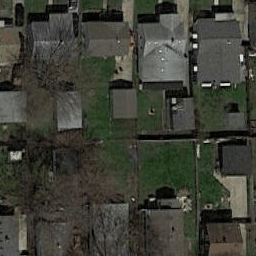} &
    \myfig{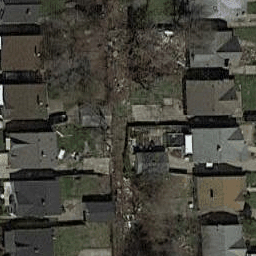} &
    \myfig{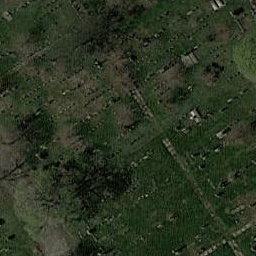} &
    {\boxcolor \fbox{\parbox[c][0.12\textwidth][c]{0.125\textwidth}{\centering \texttt{\textcolor{black}{dense}}}}} \\

    \addlinespace[2.5pt] 
    {\boxcolorim \fbox{\myfig{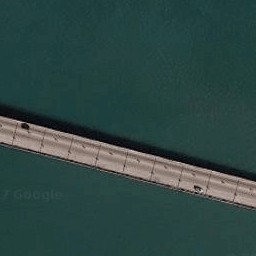}}} &
    \myfig{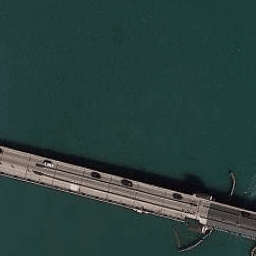} &
    \myfig{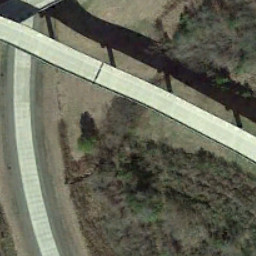} &
    \myfig{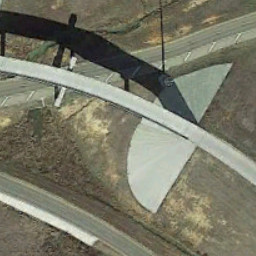} &
    \myfig{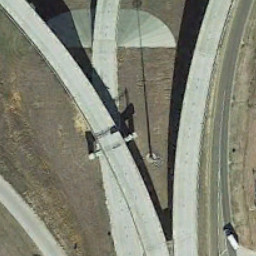} & 
    \myfig{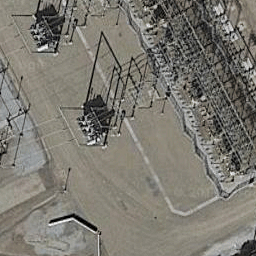} &
    {\boxcolor \fbox{ \parbox[c][0.12\textwidth][c]{0.12\textwidth}{\centering \texttt{\textcolor{black}{concrete}}}}} \\

    \mycaption{Query Image}&
    \mycaption{Image only: $\lambda = 0$} &
    \mycaption{$\lambda = 0.5$} &
    \mycaption{$\lambda = 0.75$} &
    \mycaption{$\lambda = 0.95$} &
    \mycaption{Text only: $\lambda = 1$} &
    \mycaption{Query Text}\\

\end{tabular}
\vspace{-6pt}
\captionof{figure}{We introduce remote sensing composed image retrieval (RSCIR), a novel and expressive remote sensing image retrieval (RSIR) task integrating both image and text in the search query. We also introduce \ours, a flexible, 
training-free method based on vision-language models, utilizing a weighting parameter $\lambda$ for more image- or text-oriented results, with $\lambda \to 0$ or $\lambda \to 1$ respectively. For each \textcolor{imagequerycolor}{query image} and \textcolor{textquerycolor}{query text}, retrieved images shown for different $\lambda$.}
\label{fig:teaser}
\par\vspace{10pt}
}

\maketitle
\begin{abstract}
This work introduces composed image retrieval to remote sensing. It allows to query a large image archive by image examples alternated by a textual description, enriching the descriptive power over unimodal queries, either visual or textual.  Various attributes can be modified by the textual part, such as shape, color, or context.
A novel method fusing image-to-image and text-to-image similarity is introduced.
We demonstrate that a vision-language model possesses sufficient descriptive power and no further learning step or training data are necessary.
We present a new evaluation benchmark focused on color, context, density, existence, quantity, and shape modifications. Our work not only sets the state-of-the-art for this task, but also serves as a foundational step in addressing a gap in the field of remote sensing image retrieval. Code at: \href{https://github.com/billpsomas/rscir}{https://github.com/billpsomas/rscir}.
\end{abstract}
\vspace{-5pt}
\section{Introduction}
\label{sec:intro}

In recent years, earth observation (EO) through remote sensing (RS) has witnessed an enormous growth in data volume, creating a challenge in managing and extracting relevant information. 
The capacity to efficiently organize extensive archives and quickly \emph{retrieve} specific images is crucial. 

Remote sensing image retrieval (RSIR)~\cite{agouris}, which aims to search and retrieve images from RS image archives, has emerged as a key solution. RSIR methods can be categorized into \emph{unisource} and \emph{cross-source}~\cite{zhou2023remote}, where the categorization is based on whether the query image and the retrieved images are from the same source. In the case of unisource, there exists \emph{single-label}~\cite{hou2020exploiting, wang2021learnable, sumbul2022plasticity, wang2022novel, zhao2021global, cheng2021novel} and \emph{multi-label}~\cite{kang2020graph, sumbul2021novel, sumbul2021informative, cheng2021semantic, imbriaco2021toward, shao2020multilabel} retrieval, depending on whether an image is associated with one or multiple labels respectively. In the case of cross-source, the term ``source'' is used loosely and can correspond to modality, view, etc. 

In all cases, RSIR methods encounter a major limitation: the reliance on a query of single modality. This constraint often restricts users from fully expressing their specific requirements, especially given the complex and dynamic nature of Earth's surface as depicted in RS imagery. Ideally, users would benefit from a system that allows them to articulate nuanced modifications or specifications in conjunction with an image-based query. This is where composed image retrieval (CIR)~\cite{tirg, val, lbf, combiner, cosmo, pic2word} comes into play. 
CIR, integrating both image and text in the search query, is designed to retrieve images that are not only visually similar to the \emph{query image} but also relevant to the details of the accompanying \emph{query text}.
By incorporating CIR into RS, we aim to offer a more expressive and flexible search capability that aligns closely with the needs of users in this field. 

In this paper, we recognize, present and qualitatively evaluate the capabilities and challenges that CIR introduces within the RS domain. We demonstrate how users can now pair a query image with a query text specifying modifications related to \emph{color}, \emph{context}, \emph{density}, \emph{existence}, \emph{quantity}, \emph{shape}, \emph{size} or \emph{texture} of one or more classes. Quantitatively, we focus on color, context, density, existence, quantity, and shape modifications, establishing a benchmark and an evaluation protocol. Our approach is training-free by using a frozen vision-language model. 

In summary, we make the following contributions:
\begin{enumerate}[itemsep=2pt, parsep=0pt, topsep=3pt]
    \item We are the first to introduce composed image retrieval into remote sensing, accompanied with \dataset, a benchmark dataset. 
    \item We introduce \ours, a training-free method utilizing a modality control parameter for more image- or text-oriented results according to the needs of each search, as shown in \autoref{fig:teaser}.
    \item We evaluate both qualitatively and quantitatively the performance of \ours, setting the state-of-the-art on remote sensing composed image retrieval.
\end{enumerate}



\vspace{-5pt}
\section{Related Work}
\label{sec:related}

\hspace{1.5em}\textbf{Remote Sensing Image Retrieval} 
With the aim to effectively \emph{search} and \emph{retrieve} information from extensive RS image archives, remote sensing image retrieval (RSIR) can be categorized into \emph{unisource} and \emph{cross-source}~\cite{zhou2023remote}.
Initially, RSIR methods focus on handcrafted and low-level visual features~\cite{mamatha2010content, ma2014improved, piedra2013fuzzy, li2004integrated, bhagavathy2006modeling, wang2012remote, shao2014remote, wang2013remote, chaudhuri2017multilabel, dai2017novel}. With the advent of deep learning, neural networks are utilized for unisource \emph{single-label} retrieval: (a) as feature extractors~\cite{li2016content, hu2016delving, boualleg2018enhanced, ye2018remote, napoletano2018visual, ge2018exploiting, tang2018unsupervised, imbriaco2019aggregated, sadeghi2019scalable, hou2020exploiting}, (b) trained from scratch~\cite{zhou2017learning, zhang2021triplet, zhuo2021remote, liu2020remote, wang2021learnable, sumbul2022plasticity, wang2016three, chaudhuri2019siamese}, (c) integrating attention modules~\cite{wang2022novel, wang2020attention, xiong2019discriminative, chaudhuri2021attention} and (d) using metric learning~\cite{zhao2021global, cao2020enhancing, cheng2021novel, liu2020eagle, fan2020global, liu2020similarity}. Neural networks are also used for unisource \emph{multi-label}~\cite{chaudhuri2017multilabel, kang2020graph, sumbul2021novel, sumbul2021informative, cheng2021semantic, imbriaco2021toward, shao2020multilabel}, cross-source \emph{cross-sensors}~\cite{li2018learning, ma2021cross, xiong2020discriminative, xiong2020learning}, cross-source \emph{cross-modal}~\cite{chaudhuri2021attention, xu2020mental, sun2021multisensor, sumbul2021bigearthnet, lv2021fusion, yuan2022remote, yuan2022exploring} and cross-source \emph{cross-view} retrieval~\cite{hu2018cvm, zeng2022geo, tian2020cross, lin2015learning, khurshid2019cross, shi2022beyond}. Our work fills a notable gap and enhances user intent expression in RSIR by combining query image with query text. 

\textbf{Composed Image Retrieval} Image-to-image~\cite{rtc19, gar+16, nas+17} and text-to-image~\cite{sarafianos2019adversarial, zhang2020context, devise} retrieval provide ways to explore large image archives. However, the most accurate and flexible way to express the user intent is a query \emph{composed} of both an image and a text. Composed Image Retrieval (CIR)~\cite{tirg, val, lbf, combiner, cosmo, pic2word} aims to retrieve images not only visually similar to the query image, but also altered to align with the specifics of the query text. Traditionally, CIR methods are supervised by \emph{triplets} of the form \emph{query image, query text, target image}~\cite{tirg, jvsm, yin2020disentangled, lbf, rhg+15, val, cosmo, artemis}. The labor-intensive process of labeling such triplets limit early works to specific applications in fashion~\cite{fashion200k, shoes, fashioniq}, physical states~\cite{mit_states}, object attributes and composition~\cite{tirg, lmb+14, mpc}. The emergence of vision-language models (VLMs)~\cite{clip, align, blip} led to their integration into CIR, introducing \emph{zero-shot composed image retrieval} (ZS-CIR)~\cite{pic2word, searle, cirevl}. This increases the spectrum of possible applications~\cite{searle}. Methods are trained using unlabeled images~\cite{searle, pic2word}, or are not trained at all~\cite{cirevl}. Recognizing the unexplored potential of CIR in RS, our work pioneers its introduction in this domain, particularly leveraging ZS-CIR empowered by VLMs.

\textbf{Vision-Language Models} The emergence of vision-language models (VLMs)~\cite{clip, align, blip, blip2} revolutionizes the field of multimodal learning. Trained on large-scale datasets~\cite{laion}, these models map images and text into a shared embedding space. Apart from zero-shot classification, CLIP~\cite{clip} can be used for detection~\cite{gu2021open}, segmentation~\cite{liang2023open} and captioning~\cite{tewel2022zerocap}. CLIP can also be aligned to be used with medical data~\cite{wang2022medclip} or satellite data~\cite{liu2023remoteclip, klemmer2023satclip}. In this work, we leverage CLIP and RemoteCLIP~\cite{liu2023remoteclip}, a vision-language model for remote sensing, in a training-free setting.
\vspace{-5pt}
\section{Method}
\label{sec:method}

\begin{figure*}[t]
\centering
\vspace{-10pt}
\includegraphics[trim={0cm 0cm 0cm 0cm},width=1.0\linewidth]{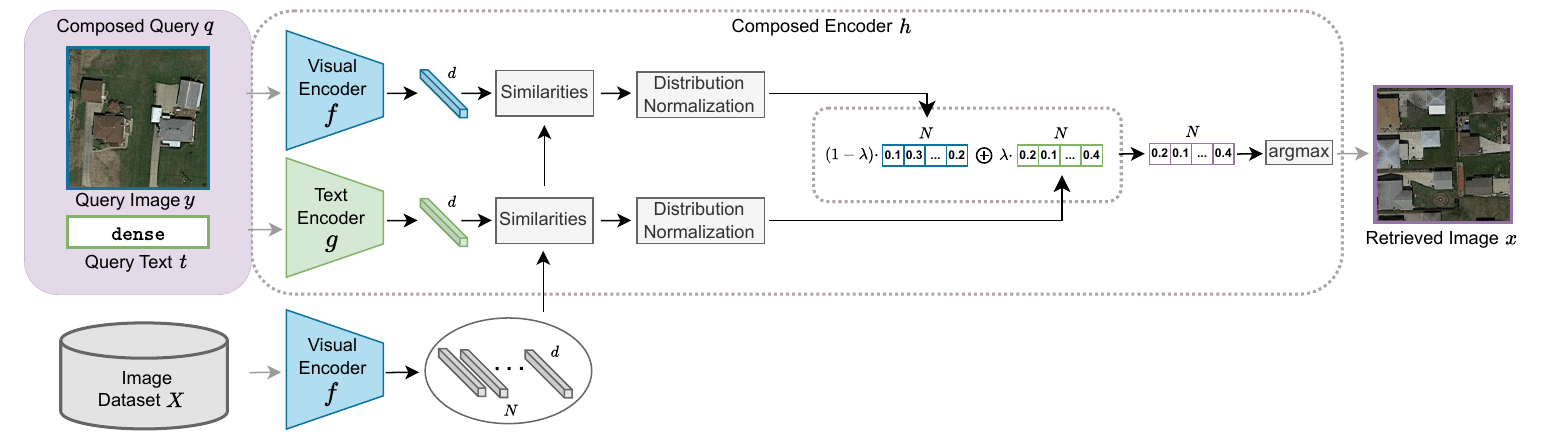}
\caption{
\emph{\ours: A \textsc{Wei}ghted \textsc{Com}posed Image Retrieval Method.} It utilizes a dual-encoder approach to process both \textcolor{imagequerycolor}{query image $y$} and \textcolor{textquerycolor}{query text $t$}. Initially, the \textcolor{imagequerycolor}{query image} is passed into a visual encoder $f$ and the \textcolor{textquerycolor}{query text} into a text encoder $g$, producing corresponding $d$-dimensional representations. Subsequently, similarity scores with the representations in the image dataset are calculated. These scores are then normalized and combined using a convex combination controlled by a $\lambda \in [0,1]$. Finally, an argmax(argsort) operation identifies the most relevant 
\textcolor{retrievedimagecolor}{retrieved image(s) $x$.}}
\label{fig:method}
\vspace{-10pt}
\end{figure*}

\subsection{Problem formulation}
\label{sec:prelim}

In composed image retrieval, the goal is to retrieve images based on a \emph{composed image-text query}, that is, a query that consists of a \emph{visual} part, the query image, and a \emph{textual} part, the query text. In this work, we introduce remote sensing composed image retrieval. To do so, we establish a benchmark and an evaluation protocol.





We denote the query image as $y$, its class as $C_y$ and an attribute of the depicted class as $A_y$. We also denote the query text as t, which represents a modified target attribute $A_t$. We refer to the two queries as the composed query, $q = (y, t)$. Given an image dataset $X$, our goal is to retrieve images from $X$ that share class with the query image class $C_y$ and have the attribute $A_t$ defined by the text query $t$. Retrieval aims to rank images $x \in X$ with respect to their composed similarity $s(q, x) \in R$ to the query. The task is extendable to multiple classes and multiple attributes.


To define $s$, we make use of pre-trained VLMs that consist of a \emph{visual encoder} \(f: \cI \to \mathbb{R}^d\) and a \emph{text encoder} \(g: \cT \to \mathbb{R}^d\), which map input images from image space \(\cI\) and words from the text space \(\cT\) to the same embedding space with dimension \(d\). We extract the visual embedding \(\mathbf{v}_y = f(y) \in \mathbb{R}^d\) and the text embedding \(\mathbf{v}_t = g(t) \in \mathbb{R}^d\) to use as queries. Finally, the embedding of a dataset image \(x \in X\) is denoted as \(\mathbf{v}_x = f(x) \in \mathbb{R}^d\). All embeddings are \(\ell_2\)-normalized.

\subsection{Baselines}
\label{sec:base}

\hspace{1.5em}\textbf{Unimodal} baselines rely solely on a single type of query to determine similarity.
We denote: \emph{text-only} by \(s_g(q, x) = g(t)^Tf(x)\) and \emph{image-only} by \(s_f(q, x) = f(y)^Tf(x)\). Unimodal baselines are expected to fail since the final similarity cannot embody information from both image and text.

\textbf{Multimodal} combines the two unimodal approaches by averaging their similarities:

\begin{equation}
s_a(q, x) = \frac{s_g(q, x) + s_f(q, x)}{2}
\vspace{0.3em}
\end{equation}

Note that this baseline is equivalent to averaging the two features $g(t), f(y)$ and then calculating the similarities once. The drawback of this approach is that the features that come from same modalities have similarities significantly greater than the cross-modal similarities, making it an approach biased in favor of the image query.

\subsection{\ours}
\label{sec:base}


In our proposed method, \ours, we estimate the similarities of the image query $s_f(q, x)$ and the text query $s_g(q, x)$ with the database. Then we perform similarity normalization in order to have a starting point of equal contribution from both modalities and we notate $s_f'(q, x), s_g'(q, x)$. Finally, we use the weighted average of the two similarity sets using a modality control parameter $\lambda$:

\begin{equation}
s_{WC}(q, x) = \lambda s_g'(q, x) + (1-\lambda) s_f'(q, x)
\vspace{0.3em}
\end{equation}

\textbf{Similarity Normalization} In order to ensure that both image and text queries contribute equally to the retrieval, we normalize their similarities with the database. We first transform the empirical distribution of similarity scores into a standard normal distribution. 
Subsequently, we apply the cumulative distribution function (CDF) of the standard normal distribution to the standardized data, resulting in values that range between 0 and 1. Assuming the standardized data adhere to a normal distribution, this transformation yields data that approximates a uniform distribution. Transforming data into a uniform distribution diminishes the influence of outliers and reduces skewness, smoothing any excessively peaked distributions. This approach leads to more robust similarity scores.

\textbf{The modality control parameter $\lambda$} After normalizing the similarities, we can control the influence of each modality using a parameter $\lambda$ as a weight. Here $\lambda=0$ refers to image-only retrieval, $\lambda=1$ to text-only retrieval and $\lambda=0.5$ to equal contribution of image and text, as shown in \autoref{fig:teaser}. The full \ours method is summarized in~\autoref{fig:method}.

\vspace{-5pt}
\section{Experiments}
\label{sec:exp}

\subsection{Datasets, networks and evaluation protocol}
\label{sec:data}


\hspace{1.5em}\textbf{Datasets} To evaluate quantitatively the methods, we introduce \dataset, a new benchmark based on PatternNet~\cite{zhou2018patternnet}. PatternNet is a large-scale high-resolution remote sensing image retrieval dataset. There are 38 classes and each class has 800 images of size 256×256 pixels. In \dataset, we select some classes to be depicted in query images, and add a query text that defines an attribute relevant to that class. 
For instance, query images of ``swimming pools'' are combined with text queries defining ``shape'' as ``rectangular'', ``oval'', and ``kidney-shaped''. In total, \dataset includes six attributes consisted of up to four different classes each. Each attribute can be associated with two to five values per class. The number of positives ranges from 2 to 1345 and there are more than 21k queries in total. Statistics for two out of six attributes are shown in~\autoref{tab:dataset}. 


\begin{table}[htbp]
\scriptsize
\renewcommand\theadfont{\scriptsize}
\centering
\begin{tabular}{ccccc}
\toprule
\Th{Attribute} &
\Th{Class}  &
\Th{Value} &
\Th{\#Positives}    &
\Th{\#Queries}   \\
\midrule

\multirow{11}{*}{color} & \multirow{2}{*}{airplane} & white & 672 & 53 \\ & & purple & 53 & 672 \\ & \multirow{2}{*}{nursing home} & white & 85 & 383 \\ & & gray & 383 & 85 \\ 
& \multirow{2}{*}{crosswalk} & white & 412 & 388 \\
& & yellow & 388 & 412 \\
& \multirow{5}{*}{tennis court} & blue & 339 & 287 \\
& & brown & 2 & 624 \\
& & gray & 50 & 576 \\
& & green & 211 & 415 \\
& & red & 24 & 602 \\ \midrule
\multirow{7}{*}{shape} & \multirow{3}{*}{swimming pool} & rectangular & 261 & 299 \\ 
& & oval & 52 & 508 \\ 
& & kidney-shaped & 247 & 313 \\ 
& \multirow{2}{*}{river} & curved & 177 & 623 \\
& & straight & 623 & 177 \\
& \multirow{2}{*}{road} & cross & 800 & 800 \\
& & round & 800 & 800 \\

\bottomrule
\end{tabular}
\vspace{-5pt}
\caption{\emph{Statistics for color and shape attributes of} \dataset, the first RSCIR benchmark.}
\vspace{-10pt}
\label{tab:dataset}
\end{table}

\textbf{Networks} We use the pre-trained CLIP~\cite{clip} and RemoteCLIP~\cite{liu2023remoteclip}, both with a ViT-L/14 image encoder.

\textbf{Evaluation Protocol} We evaluate using mAP. Average Precision (AP) is the average of the precision values obtained for the set of top-$k$ results, up to each relevant item found in the ranking. The mAP is then the mean of these AP values over all queries. 

\begin{figure}[ht]
\vspace{-10pt}
\centering 
\begin{minipage}[b]{0.45\linewidth}
\centering
\footnotesize
\setlength{\tabcolsep}{1pt}
{
\footnotesize
\centering
\newcommand{\myg}[1]{\includegraphics[width=0.31\textwidth,valign=c]{#1}}

\setlength{\tabcolsep}{1pt}
\begin{tabular}{@{}ccc@{}}

    \mycaption{Query} & 
    \mycaption{Query} & 
    \mycaption{Retrieved} \\

    \mycaption{Image} & 
    \mycaption{Text} & 
    \mycaption{Image} \\
    
    \myg{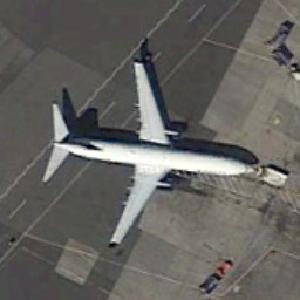} & 
    \fbox{\parbox[c][0.26\textwidth][c]{0.26\textwidth}{\centering \texttt{purple}}} & 
    \myg{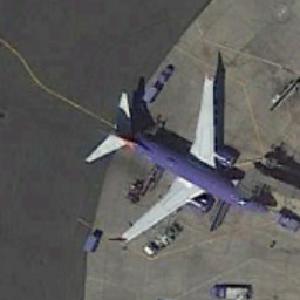} \\
    \mc{3}{(a) Color}\\

    \mc{3}{\vspace{-2.1ex}}\\

    \myg{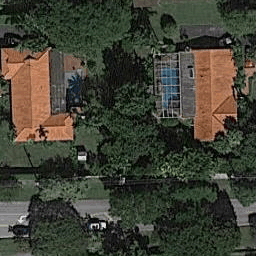} & 
    \fbox{\parbox[c][0.26\textwidth][c]{0.26\textwidth}{\centering \texttt{dense}}} & 
    \myg{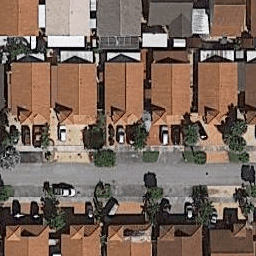} \\

    \mc{3}{(c) Density
    }\\ 

    \mc{3}{\vspace{-2.1ex}}\\

    \myg{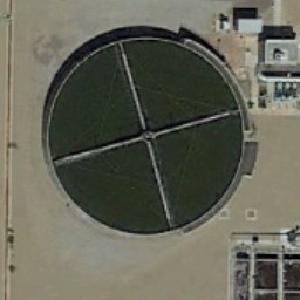} & 
    \fbox{\parbox[c][0.26\textwidth][c]{0.26\textwidth}{\centering \texttt{four}}} & 
    \myg{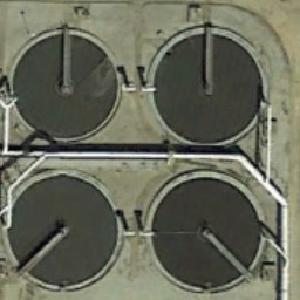} \\

    \mc{3}{(e) Quantity
    }\\ 

    \mc{3}{\vspace{-2.1ex}}\\

    \myg{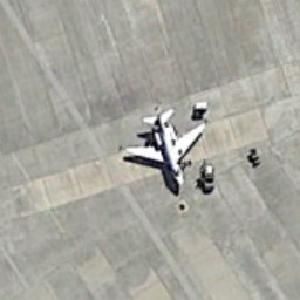} & 
    \fbox{\parbox[c][0.26\textwidth][c]{0.26\textwidth}{\centering \texttt{big}}} & 
    \myg{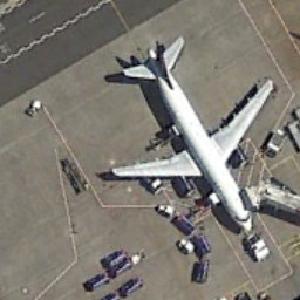} \\

    \mc{3}{(g) Size
    }\\ 
    
\end{tabular}
}
\end{minipage}
\hfill 
\begin{minipage}[b]{0.45\linewidth}
\centering
\footnotesize
\setlength{\tabcolsep}{1pt}
{
\scriptsize
\footnotesize
\centering
\newcommand{\myg}[1]{\includegraphics[width=0.31\textwidth,valign=c]{#1}}

\setlength{\tabcolsep}{1pt}
\begin{tabular}{@{}ccc@{}}
   
    \mycaption{Query} & 
    \mycaption{Query} & 
    \mycaption{Retrieved} \\

    \mycaption{Image} & 
    \mycaption{Text} & 
    \mycaption{Image} \\

    \mc{3}{\vspace{-2.1ex}}\\

    \myg{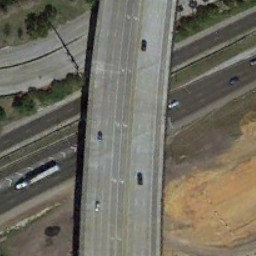} & 
    \fbox{\parbox[c][0.26\textwidth][c]{0.26\textwidth}{\centering \texttt{water}}} & 
    \myg{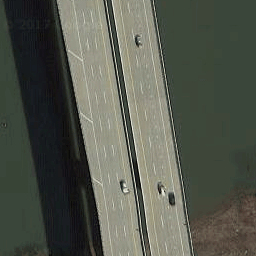} \\

    \mc{3}{(b) Context
    }\\ 

    \mc{3}{\vspace{-2.1ex}}\\

    \myg{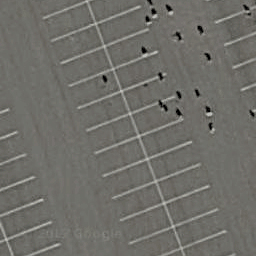} & 
    \fbox{\parbox[c][0.26\textwidth][c]{0.26\textwidth}{\centering \texttt{full}}} & 
    \myg{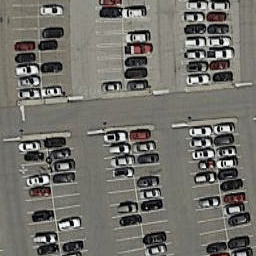} \\

    \mc{3}{(d) Existence
    }\\ 

    \mc{3}{\vspace{-2.1ex}}\\
    
    \myg{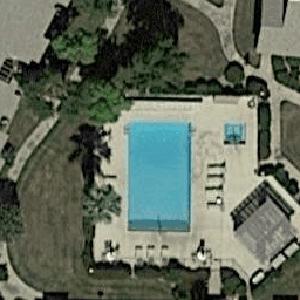} & 
    \fbox{\parbox[c][0.26\textwidth][c]{0.26\textwidth}{\centering \texttt{oval}}} & 
    \myg{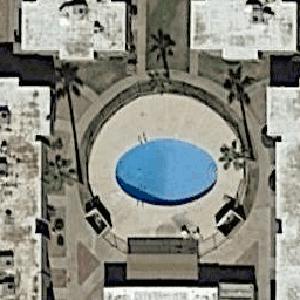} \\

    \mc{3}{(f) Shape
    }\\

    \mc{3}{\vspace{-2.1ex}}\\

    \myg{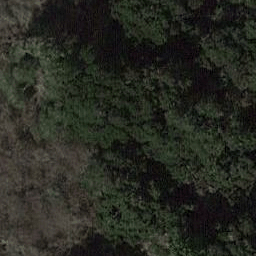} & 
    \fbox{\parbox[c][0.26\textwidth][c]{0.26\textwidth}{\centering \texttt{fine}}} & 
    \myg{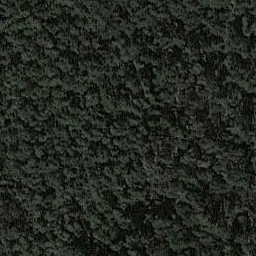} \\

    \mc{3}{(h) Texture
    }\\ 

\end{tabular}
}
\end{minipage}
    \vspace{-10pt}
    \caption{\emph{Demonstrating remote sensing composed image retrieval}. Subfigures (a) to (h) depict the key attributes: color, context, density, existence, quantity, shape, size, and texture. Each one illustrates various utilizations of composed image retrieval in remote sensing. Subfigures (b), (d) are examples that extend the task to multiple classes and attributes.
    }
    \label{fig:presentation}
    \vspace{-10pt}
\end{figure}

\subsection{Experimental results}
\label{sec:results}

\hspace{1.5em}\textbf{Qualitative results} In~\autoref{fig:presentation}, we present the qualitative results of performing composed image retrieval in \dataset using \ours with RemoteCLIP. Each example corresponds to one of the selected attributes with the query text specifying a modification in each attribute value. 

\textbf{Comparison with baselines} As shown in~\autoref{tab:sota}, \ours outperfoms both unimodal (``Text", ``Image") and multimodal (``Text \& Image") baselines by a large margin. In particular, it outperforms the second best by 8.95\% mAP using CLIP and 15.14\% mAP using RemoteCLIP on average. 

\begin{table}[htbp]
\vspace{-10pt}
\centering
\begin{subtable}{\linewidth}
\centering
\scriptsize
\setlength{\tabcolsep}{1pt}
\caption{CLIP~\cite{clip}}
\begin{tabular}{lccccccc}
\toprule
\Th{Method} &
\Th{Color}  &
\Th{Context} &
\Th{Density}    &
\Th{Existence}   &
\Th{Quantity}   &
\Th{Shape}   &
\Th{Avg} \\ \midrule 
Text            & 13.47      & 4.83      & 3.58      & 4.38       & 3.31      & 6.22      & 5.97 \\
Image           & 14.66      & 8.32      & 13.49     & 13.50      & 7.84      & 15.76     & 12.26 \\
Text \& Image   & 23.13      & 11.02     & 15.87     & \textbf{13.77}      & 10.13      & 21.38     & 15.88  \\
\midrule 
\textbf{$\ours_{\lambda=0.5}$}  & 46.08 & 17.45 & 16.49   & 9.24   & 18.15 & 23.97 & 21.90 \\
\rowcolor{LightSteelBlue1}
\textbf{$\ours_{\lambda=0.3}$} & \textbf{46.74} & \textbf{20.97} & \textbf{22.07}   & 12.07   & \textbf{20.96} & \textbf{26.22} & \textbf{24.83} \\
\bottomrule
\end{tabular}
\end{subtable}

\vspace{0.2em} 

\begin{subtable}{\linewidth}
\centering
\scriptsize
\setlength{\tabcolsep}{1pt}
\caption{RemoteCLIP~\cite{liu2023remoteclip}}
\begin{tabular}{lccccccc}
\toprule
\Th{Method} &
\Th{Color}  &
\Th{Context} &
\Th{Density}    &
\Th{Existence}   &
\Th{Quantity}   &
\Th{Shape}   &
\Th{Avg} \\ \midrule 
Text            & 10.75    & 8.87          & 22.16       & 12.49     & 8.25  & 24.12  & 14.44        \\
Image           & 14.40    & 6.62          & 15.11       & 9.29     & 6.99   & 15.18  & 11.27       \\
Text \& Image   & 23.67    & 10.01         & 18.45       & 10.56     & 7.97   & 19.63  & 15.05         \\
\midrule
\textbf{$\ours_{\lambda=0.5}$}  & \textbf{43.68} & 31.45 & 39.94   & 14.27   & 20.51    & 29.78    & 29.94 \\
\rowcolor{LightSteelBlue1}
\textbf{$\ours_{\lambda=0.6}$}  & 41.04 & \textbf{31.59} & \textbf{41.56}   & \textbf{14.79}   & \textbf{20.79}   & \textbf{31.24}   & \textbf{30.19} \\
\bottomrule
\end{tabular}
\end{subtable}
\vspace{-5pt}
\caption{\emph{Attribute modification mAP (\%)} on \dataset using CLIP (a) and RemoteCLIP (b); comparison of \ours with baselines. For each attribute value of an attribute (e.g. ``rectangular" of \Th{Shape}), average mAP over all the rest attribute values (e.g. ``oval" of \Th{Shape}). \Th{AVG}: average mAP over all combinations.}
\vspace{-10pt}
\label{tab:sota}
\end{table}

\subsection{Ablation study}
\label{sec:ablation}

\hspace{1.5em}\textbf{The impact of $\lambda$} In~\autoref{tab:ablation} we show the impact of modality control parameter $\lambda$ on \ours using RemoteCLIP. $\lambda=0$ refers to image-only, $\lambda=1$ to text-only retrieval. For $\lambda=0.6$ we get the best average mAP, thus we set this as our method's default. The same study for CLIP gives $\lambda=0.3$.

\begin{table}[htbp]
\vspace{-5pt}
\centering
\scriptsize
\resizebox{1\columnwidth}{!}{%
\begin{tabular}{lccccccccccc}
\toprule
$\lambda$ & 0 & 0.1 & 0.2 & 0.3 & 0.4 & 0.5 & 0.6 & 0.7 & 0.8 & 0.9 & 1.0 \\
\midrule
Color     & \gradientcolor{14.5} & \gradientcolor{55.3} & \gradientcolor{53.0} & \gradientcolor{49.6} & \gradientcolor{46.4} & \gradientcolor{43.7} & \gradientcolor{41.0} & \gradientcolor{38.2} & \gradientcolor{35.0} & \gradientcolor{30.4} & \gradientcolor{10.8} \\
Context     & \gradientcontext{6.6}     & \gradientcontext{13.3}     & \gradientcontext{20.2}     & \gradientcontext{25.7}     & \gradientcontext{29.5}     & \gradientcontext{31.5}     & \gradientcontext{31.6}     & \gradientcontext{29.6}     & \gradientcontext{24.8}     & \gradientcontext{16.9}     & \gradientcontext{8.9} \\
Density     & \gradientdensity{15.1}     & \gradientdensity{23.3}     & \gradientdensity{29.5}     & \gradientdensity{34.0}     & \gradientdensity{37.4}     & \gradientdensity{39.9}     & \gradientdensity{41.6}     & \gradientdensity{42.0}     & \gradientdensity{40.7}     & \gradientdensity{35.9}     & \gradientdensity{22.2} \\
Existence     & \gradientexistence{9.3}     & \gradientexistence{10.3}     & \gradientexistence{11.1}     & \gradientexistence{12.3}     & \gradientexistence{13.5}     & \gradientexistence{14.3}     & \gradientexistence{14.8}     & \gradientexistence{15.0}     & \gradientexistence{14.8}     & \gradientexistence{14.0}     & \gradientexistence{12.5} \\
Quantity     & \gradientquantity{7.0}     & \gradientquantity{17.6}     & \gradientquantity{18.9}    & \gradientquantity{19.7}     & \gradientquantity{20.2}     & \gradientquantity{20.5}     & \gradientquantity{20.8}     & \gradientquantity{20.9}     & \gradientquantity{20.8}     & \gradientquantity{20.1}     & \gradientquantity{8.3} \\
Shape     & \gradientshape{15.2}     & \gradientshape{23.8}     & \gradientshape{24.7}     & \gradientshape{26.2}     & \gradientshape{28.0}     & \gradientshape{29.8}     & \gradientshape{31.2}     & \gradientshape{32.0}     & \gradientshape{32.0}     & \gradientshape{31.3}     & \gradientshape{24.1} \\
Average     & \gradientaverage{11.3}     & \gradientaverage{23.9}     & \gradientaverage{26.2}     & \gradientaverage{27.9}     & \gradientaverage{29.2}     & \gradientaverage{29.9}     & \gradientaverage{30.2}     & \gradientaverage{29.6}     & \gradientaverage{28.0}     & \gradientaverage{24.8}     & \gradientaverage{14.4} \\

\bottomrule
\end{tabular}
}
\vspace{-5pt}
\caption{\emph{The effect of the modality control parameter $\lambda$ on} \ours using RemoteCLIP, measured in attribute modification mAP.}
\vspace{-25pt}
\label{tab:ablation}
\end{table}

\vspace{-5pt}
\section{Conclusions}
\label{sec:conclusions}

We introduce remote sensing composed image retrieval, a novel task integrating both image and text in the search query, accompanied with \dataset, a benchmark dataset. We  demonstrate its versatility through use cases modifying attributes like color or shape and also introduce \ours, a flexible and training-free method utilizing a modality control parameter $\lambda$, setting the state-of-the-art on the task.

\vspace{5pt}
{\small 
\noindent \textbf{Acknowledgements}
Bill was supported by the RAMONES H2020 project (grant: 101017808). This work was also supported by the HFRI under the BiCUBES project (grant: 03943), by the Czech Technical University in Prague grant No. SGS23/173/OHK3/3T/13, the Junior Star GACR GM 21-28830M, and the CTU institutional support (Future fund). NTUA thanks NVIDIA for the donation of GPU hardware.
}

\bibliographystyle{IEEEbib}
\bibliography{refs}

\end{document}